\documentclass[preprint,12pt]{elsarticle}
\pdfoutput=1
\usepackage{hyperref}

\usepackage{framed}

\usepackage{amssymb}
\usepackage{amsmath}
\usepackage{enumitem}
\usepackage{graphicx}
\usepackage{url}
\usepackage{listings}
\usepackage{bm}
\usepackage{multirow}
\usepackage{array}
\usepackage{color}
\usepackage{threeparttable}
\usepackage{float}
\usepackage{wrapfig}
\usepackage{xcolor}
\usepackage{tikz}
\usepackage{makecell}
\usepackage{cuted}
\usepackage{capt-of}
\definecolor{shadecolor}{rgb}{0.95,0.95,0.92}
\usepackage{xcolor}
\usepackage{xcolor,colortbl}
\usepackage{listings}
\usepackage{ulem}

\usepackage{pifont}
\newcommand{\cmark}{\ding{51}}%
\newcommand{\xmark}{\ding{55}}%

\definecolor{shadecolor}{rgb}{0.95,0.95,0.92}
\definecolor{codegreen}{rgb}{0,0.6,0}
\definecolor{codegray}{rgb}{0.5,0.5,0.5}
\definecolor{codepurple}{rgb}{0.58,0,0.82}
\definecolor{backcolour}{rgb}{0.95,0.95,0.92}

\lstdefinestyle{mystyle}{
    commentstyle=\color{codegreen},
    keywordstyle=\color{magenta},
    numberstyle=\tiny\color{codegray},
    stringstyle=\color{codepurple},
    basicstyle=\ttfamily\footnotesize,
    breakatwhitespace=false,
    breaklines=true,
    captionpos=b,
    keepspaces=true,
    numbers=left,
    numbersep=5pt,
    showspaces=false,
    showstringspaces=false,
    showtabs=false,
    tabsize=2
}

\usetikzlibrary{shapes,arrows,arrows,positioning,matrix}
\urldef{\mailsc}\path|matliuj@nus.edu.sg|

\journal{\LaTeX}

\begin{document}

\begin{frontmatter}

\title{Dim the Lights! $-$  Low-Rank Prior Temporal Data  \\ for Specular-Free Video Recovery}


\author[1,2]{Samar M. Alsaleh}
\author[3]{Angelica I. Aviles-Rivero}
\author[4]{No\'{e}mie Debroux}
\author[1]{ and \\James K. Hahn}
\address[1]{Department of Computer Science, George Washington
University, Washington DC, U.S.A. {\tt\small \{sm57,hahn\}@gwu.edu,}}
\address[2]{Department of Computer Science, Taibah University, Saudi Arabia}
\address[3]{Department of Pure Mathematics and Mathematical Statistics (DPMMS), University of Cambridge, U.K.    {\tt\small ai323@cam.ac.uk} }
\address[4]{Universit\'e Clermont Auvergne, CNRS, SIGMA Clermont, Institut Pascal, F-63000 Clermont-Ferrand, France {\tt\small noemie.debroux@uca.fr}}

\begin{abstract}
The appearance of an object is significantly affected by the illumination conditions in the environment. This is more evident with strong reflective objects as they suffer from more dominant specular reflections, causing information loss and discontinuity in the image domain. In this paper, we present a novel framework for specular-free video recovery with special emphasis on dealing with complex motions coming from objects or camera. Our solution is a two-step approach that allows for both detection and restoration of the damaged regions on video data. We first propose a spatially adaptive detection term that searches for the damage areas. We then introduce a variational solution for specular-free video recovery that allows exploiting spatio-temporal correlations by representing prior data in a low-rank form. We demonstrate that our solution prevents major drawbacks of existing approaches while improving the performance in both detection accuracy and inpainting quality. Finally, we show that our approach can be applied to other problems such as object removal.
\end{abstract}


\end{frontmatter}

\section{Introduction}
In the science of vision, light is what enables our biological vision system to see our surroundings and identify different objects, regardless of the many unpredictable changes in the realistic environment including lighting conditions. This adaptation, however, does not extend to computer vision systems as they still struggle to robustly process illumination changes as humans do and counter for their side-effects. This is more evidenced with objects that have strong reflectivity as light changes result in more dominant specular reflections which cause information loss and discontinuity in the image domain~\cite{Tan::04, Yang::15}.

Images and videos captured in the real world are expected to have specular reflections due to the inhomogeneous nature of many materials such as plastic, metals and human skin~\cite{Tan::04}. Based on the dichromatic reflection model \cite{Shafer::85}, those materials, and many others of natural objects, tend to have two reflection components: diffuse and specular. These reflection components are formed by different physical light-surface interactions. Diffuse reflection is caused by the subsurface scattering of light, at many angles, and is a direct representation of the shape of an object. Specular reflection, on the other hand, only appears at some locations on an object’s surface and exhibits less scattering causing it to have strong intensity~\cite{Umeyama::04}. This reflection component depends heavily on the local orientation and degree of roughness of the surface and, therefore, captures important scene information, such as surface shape and light source characteristics~\cite{Fleming::04, Yang::15}.

At the practical level, the behavior and presence of specular reflections often cause significant inaccuracies or even failure of common vision algorithms such as segmentation~\cite{Deng::99, Yang::10}, visual recognition ~\cite{Chen::06, Li::17}, stereo matching~\cite{Heo::11}, medical image analysis~\cite{Kong::16,aviles2017towards}, tracking~\cite{Zhang::06}, and scene reconstruction~\cite{Furukawa::10}. For all these reasons, restoration of specular reflection has become crucial to the practicality, accuracy and robustness of computer vision systems. The problem of dealing with specular artifacts can be addressed using either single image or a video sequence. Although both perspectives have been explored by the community, the question of - \textbf{\it{how to exploit, in an efficient manner, the spatio-temporal correlations for complex sequences while keeping low computational complexity?}} still remains open and therefore it is worthy of exploration.


\medskip
\textbf{Contributions}. We present a novel framework for specular-free video recovery that is equally effective for both static and moving cameras and with the presence of object motion. Our  two-step solution is recast as an optimization problem and allows for fully-automatic detection and restoration of specular regions on video data. Our main contributions are:

\begin{itemize}
    \item We propose a computationally tractable solution based on two main components:
        \begin{itemize}\itemsep0em
            \item A low computational yet accurate detection approach based on a set of adaptive conditions.
            \item A variational framework that exploits spatio-temporal correlations in low-rank representation.
        \end{itemize}
    \item We validate exhaustively our approach, using different complex scenes. We show that our approach achieves a better approximation of the damage area than the body of literature techniques.
\end{itemize}

\section{Related Work}
The problem of specular-free video recovery can be broken down into two sub-problems: (i) damaged regions detection and (ii) missing information recovery. In this section, we review the body of literature on both sub-problems.

\medskip
\textbf{Specularity Detection}.
Generally speaking, the reflection removal problem can be seen as the problem of separating two linearly superposed components into two intrinsic images - a diffuse and a reflection image. Different works in the body of literature have been reported to solve this problem in which solutions can rely on using either multiple or single images. The former relies on the use of images of the same scene taken under different lighting~\cite{Lin::02}, from different viewpoints~\cite{Lin::02} or utilizing an additional polarizing filter~\cite{Nayar::97}. Nevertheless, the necessity of having multiple images with specific varying conditions, or of having specific hardware assistance, limits their applicability to general cases. The later overcomes this by utilizing a single image and relying on neighborhood~\cite{Tan::05,Mallick::05, Mallick::06, Shen::08, Akashi::14} or color space~\cite{Ortiz::06,Kim::13, Nguyen::14, Yang::15, Yamamoto::17} analysis and propagation. Those approaches however cannot handle large highlight areas and might lead to false specularity detection.


\medskip
\textbf{Multiple-Image/Video Restoration}.
Whether it is an image or a video, inpainting and restoration of damage areas is an ill-posed inverse problem that has no well-defined unique solution~\cite{Guillemot::14}.
Although the vast amount of redundant information that exists in multiple-images and video sequences is advantageous to the inpainting process, it presents more challenges related to computational complexity and temporal coherency.
While many literature reviews attempt to classify the solutions of video inpainting, they can be broadly viewed as variations of the exemplar-based concept originally presented in~\cite{Criminisi::04} in which missing areas are filled by propagating information from known regions. The variations come in terms of the exemplar’s shape, regular (such as patch) vs irregular (such as segment), and the exemplar’s best match search, local~\cite{Patwardhan::07, Strobel::14, Daisy::15, Huang::16} vs. global~\cite{Wexler::07,GranadosA::12, Newson::14, Ebdelli::15, Le::17}.
Although these works achieve promising results, they suffer from unpleasing artifacts and are based on a set of minimization procedures, sometimes up to five, which results in high computational demands.

\section{Specular-Free Video Recovery}
\begin{figure*}[!t]
\centering
\includegraphics[width=\textwidth]{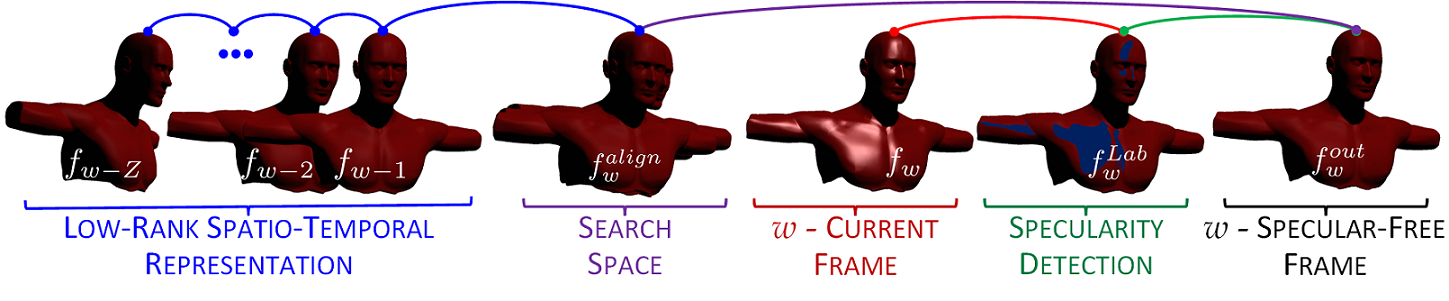}
\caption{Overview of our approach. (From left to right) prior temporal data is aligned and then used to construct a search space that, together with the result from our detection process, are used to find the optimal information for the damaged regions.}
\label{fig:feature-graphic}
\end{figure*}

In this section, we describe the two key parts of our proposed approach illustrated in Figure \ref{fig:feature-graphic}.

\subsection{Detecting Specular Reflections}
Consider an image sequence $F=\{f_w\}_{w=1}^W$, with $W$ frames where $f_w : \Omega \subset \mathbb{R}^2 \rightarrow \mathbb{R}^3$, $f_w=\{f_w^r,f_w^g,f_w^b\}$, $\Omega$ an open bounded Lipschitz subset of $\mathbb{R}^2$ of integer width $[\mathcal{A}]$, and height $[\mathcal{B}]$.
Our first stage is to detect the damaged regions and relies on the following assumptions, which appear to be realistic for complex scenes with specular reflections: \smallskip

\noindent
\textbf{(A.1)}  The reflection artifacts only appear on some locations of the image.\\
\textbf{(A.2)}  The reflections are stronger than or as intense as the target scene exhibiting less scattering. \\

We thus start by computing the amount of color dispersion at the current $w$-frame $s_w^2=\frac{1}{\mathcal{N}-1}\sum_{k=1}^{\mathcal{N}}(x_{k}^{w}-\hat{x}^{w}_{rgb})$, where $x_k^w$ and $\hat{x}_{rgb}^w$ denote the $k$-th pixel value and the mean value of the observations on the current frame $f_w$ respectively, and $\mathcal{N}=[\mathcal{A}]\times [\mathcal{B}]$ the total number of pixels in the image. We then use this information as a guideline to label pixels into two classes: specular region ($\overline{\mathcal{G}}$), i.e., damaged regions, and non-specular region ($\mathcal{G}$) such that $\Omega = \mathcal{G}\cup \overline{\mathcal{G}}$. To this end, for a given position $(i,j)\in \Omega$, the label assignation is as follows:



\begin{shaded*}
\vspace{-0.5cm}
\begin{lstlisting}[language=Python, mathescape=true]
# DETECTING SPECULAR AND NON-SPECULAR REGIONS
Set $\sigma^2(\beta^*)=\max \sigma^2(\beta)$  # compute optimal $\beta$
# COMPUTING SPECULAR REGIONS
if  $f_{w}(i,j)\in \overline{\mathcal{G}}$
    $\mathcal{I}(i, j)>\max(\mathcal{I})- s_{w}^{2} \vee \bigg(\bigg(\frac{\partial \mathcal{I}}{\partial i}\bigg)^{2}+\bigg(\frac{\partial \mathcal{I}}{\partial j}\bigg)^{2}\bigg)^{\frac{1}{2}} > \beta$
# COMPUTING NON-SPECULAR REGIONS
else
    $f_{w}(i,j)\in  \mathcal{G}$
\end{lstlisting}
\vspace{-0.5cm}
\end{shaded*}



\noindent
where $\mathcal{I}$ is the mean of the RGB channel values computed as $\mathcal{I}=\frac{1}{3}(f_w^r+f_w^g+f_w^b)$, $a \vee b$ denotes the minimum value between $a$ and $b$, and $\beta$ is optimally determined by maximizing the intra-class variance 
following the philosophy of~\cite{jenks1967data,otsu1975threshold}. It is worth noticing that for every frame $f_w$, we generate a specularity characteristic function $\phi : \Omega \rightarrow \{0,1\}$ such that $\phi(\overline{\mathcal{G}})=1$ and $\phi(\mathcal{G})=0$. For notation convenience, we will refer to $f_w^{Lab}$ for this binary map in the remainder of the text.

\subsection{Specular-Free Video Recovery}
Recently, low-rank data representation has attracted great attention in many areas including computer vision and machine learning. This is mainly because it allows keeping the relevant information in a low-dimensional space \cite{haeffele2014structured} and enables recovery of a low-rank matrix from a set of sampled entries~\cite{candes2012exact}. 
Our motivation for promoting low-rank data representation is two-fold: firstly to significantly decrease the computation time, and secondly to achieve better performance while reducing artifacts and increasing robustness to outliers. Our approach relies on this further assumption:
\smallskip

\noindent
\textbf{(A.3)} For a given image sequence $F=\{f_{w}\}_{w=1}^{W}$, we consider that there are $Z \subset F$  undamaged frames.\\
In this work, to obtain a good approximation for $Z$, we project the computed nearest neighbour patch with respect to the damage area.

\begin{figure*}
\centering
   \includegraphics[width=0.8\textwidth]{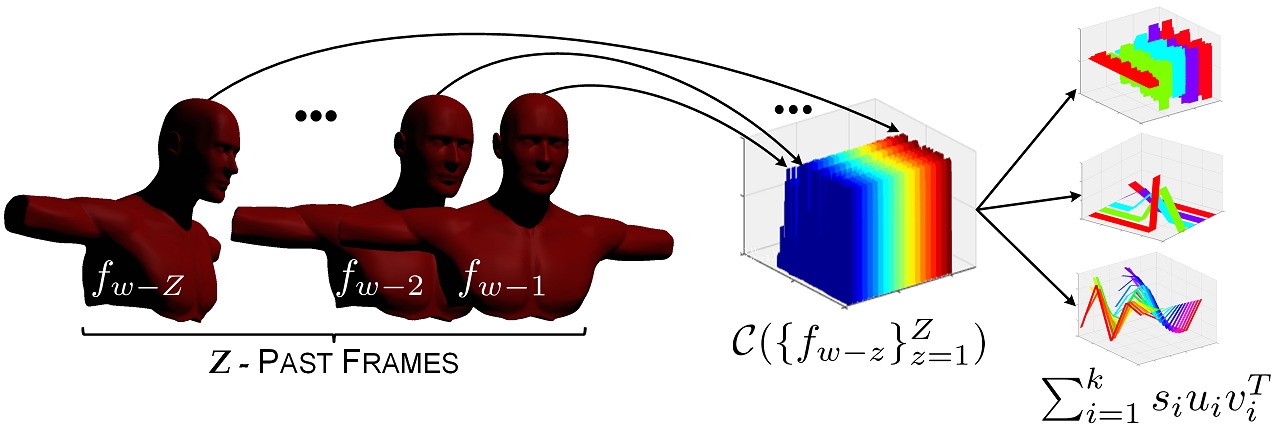}
   \caption{
Prior temporal data is arranged in a Casorati Matrix $\mathcal{C}$ in order to exploit the correlation between the data and find the low rank component $v$.   }
  \label{fig:casoratiMatrix}
\end{figure*}

Let $F_{T}^{w}=\{f_{w-z}\}_{z=1}^{Z}$ be a set of $Z$ past frames with respect to the current one $f_{w}$ where $F_{T}^{w}\subset F$. We start by constructing the following Casorati matrix $\mathcal{C}$, illustrated in Figure~\ref{fig:casoratiMatrix}, in the form:

\begin{equation}
\label{eq:Casorati}
{
\mathcal{C}=\mathcal{C}(F_{T}^{w}) = \left[
\begin{array}{c c c c}
     f_{w-1}(1,1)   &\cdots & f_{w-Z}(1,1)\\
      \vdots	 &   &\vdots \\
      f_{w-1}([\mathcal{A}], [\mathcal{B}]) &    \cdots & f_{w-Z}([\mathcal{A}], [\mathcal{B}])\\
\end{array}\right]}
\end{equation}

\noindent
for each channel color, where $ f_{w-z}(\cdot,\cdot)$ is the scalar value at a given pixel location in frame $f_{w-z}$ for the respective channel color. The idea is to exploit the strong correlation between the columns of this matrix to create a low-rank representation of the sequence (see Figure \ref{fig:casoratiMatrix}). To do so, we rely on the well-established Singular-Value Decomposition (SVD, \cite{strang05}) $\mathcal{C}=\underset{i=1}{\overset{k}{\sum}}s_iu_iv_i^T=USV^T$, with $U=(u_1,\cdots,u_{[A]})\in \mathbb{R}^{[\mathcal{A}]\times [\mathcal{A}]}$, $V=(v_1,\cdots,v_{[\mathcal{B}]})\in \mathbb{R}^{[\mathcal{B}]\times [\mathcal{B}]}$ orthogonal matrices, and $S=(s_1,\cdots,s_k,0,\cdots,0)\in \mathbb{R}^{[\mathcal{A}]\times [\mathcal{B}]}$ a diagonal matrix composed of the singular values $s_i$ with $k\leq \min([\mathcal{A}],[\mathcal{B}])$. Now, let $r\leq k$. We thus aim at solving the following problem:
\begin{equation}
    \begin{aligned}
    \hat{\mathcal{C}}=\underset{\text{rank}\,\hat{\mathcal{C}}\leq r}{\operatorname{argmin}}\,\|\mathcal{C} - \hat{\mathcal{C}}\|_F^2=\underset{i=1}{\overset{r}{\sum}}s_i u_i v_i^T,
    \end{aligned}
    \label{eq:low_rank}
\end{equation}
with $\|.\|_F$ the Frobenius matrix norm invariant to rotation and to rank. Therefore, instead of computing the SVD of a large and dense matrix, we aim at retrieving the set of $r$ dominant singular values with the associated right and left singular vectors in order to keep the most relevant information in a subspace smaller than the original one while eliminating the subspace where noise lies. This makes our method more robust to noise and outliers and reduces artifacts. Finally, using the definition of the Casoratori matrix, one is able to recover the low-rank image sequence:
\begin{equation}
\label{eq:Casorati_low_rank}
\begin{aligned}
&\hat{\mathcal{C}} = \left[
\begin{array}{c c c c}
     \bar{f}_{w-1}(1,1)   &\cdots & \bar{f}_{w-Z}(1,1)\\
      \vdots	 &   &\vdots \\
      \bar{f}_{w-1}([\mathcal{A}],[ \mathcal{B}]) &    \cdots & \bar{f}_{w-Z}([\mathcal{A}],[ \mathcal{B}])\\
\end{array}\right]=\hat{U}.\hat{S}.\hat{V}^T,\\ &\overline{F_T^w}=\{\bar{f}_{w-z}\}_{z=1}^Z,
\end{aligned}
\end{equation}
with $\hat{S}\in \mathbb{R}^{r\times r}$, $\hat{U} = \mathbb{R}^{[\mathcal{A}]\times r}$, $\hat{V}\in \mathbb{R}^{[\mathbb{B}]\times r}$. The following computations will be done using this low-rank sequence.

Our next step is to reconstruct a common search space to recover the lost information in the region $\overline{\mathcal{G}} \subset f_w^{Lab}$ using $\overline{{F}_{T}^{w}}$. This space is constructed by registering all the images in $\overline{{F}_{T}^{w}}$, that is to say by finding optimal diffeomorphic transformations characterized by the deformation fields $\varphi$ such that the images in $\overline{F_T^w}$ are aligned. This task can be cast as the following optimization problem in a variational framework, for each $z\in \{2,\cdots,Z\}$: 

\begin{equation}
\label{eq:energy}
\min_\varphi\, (\mathcal{D}(\overline{f}_{w-z+1},\overline{f}_{w-z};\varphi)+\gamma \mathcal{R}(\varphi)+ \delta \mathcal{T}(\varphi))
\end{equation}

\noindent
where $\gamma$ and $\delta$ are positive weighting parameters. $\mathcal{D}$ is a discrepancy measure motivated by robust statistics and the Huber estimator to deal with outliers and increase accuracy and robustness:
\begin{equation*}
    \begin{aligned}
   \mathcal{D}(\overline{f}_{w-z+1},\overline{f}_{w-z};\varphi) = \int_\Omega \rho(\overline{f}_{w-z+1}(\varphi(x))-\overline{f}_{w-z}(x))\,dx,
    \end{aligned}
\end{equation*}
with $\rho$ a likelihood type estimator, which is computed as:
\begin{equation}
\label{eq:estimator}
\rho(x) = \left\{ \begin{array}{c} \frac{c^2}{6}(1-(1-\frac{x}{c})^2)^3 \text{ if } |x|\leq c,\\\frac{c^2}{6} \text{ otherwise.} \end{array}\right.
\end{equation}
This particular choice of the Tukey estimator is driven by its hard rejection of outliers \cite{stewart99}.

Whilst $\mathcal{R}$ is a regularization term for the deformation field to practically solve the original highly ill-posed registration problem. The following Tikhonov penalizer $\mathcal{R}(\varphi)=\int_\Omega \|\nabla \varphi(x)\|^2\,dx$ is considered to stabilize the energy functional and to smooth the deformation field.


The last component $\mathcal{T}$ ensures physically and mechanically meaningful deformations and deals with topology preservation. This is of great importance since it guarantees the non-destruction and non-creation of structures during the registration process and controls the degree of expansions and contractions allowed. This translates as a positivity constraint on the Jacobian determinant denoted $|J_\varphi(x)|$. In this work, we use the following topology-preserving regularizer as in \cite{aviles-rivero} $\mathcal{T}(\varphi)=\int_\Omega h_\varphi(x)\,dx $, with:
\begin{equation}
\label{eq:regularizer}
h_\varphi(x) = \left\{ \begin{array}{c} e^{-|J_\varphi(x)|}+\xi \sqrt{|J_\varphi(x)|^2}\text{ if } |J_\varphi(x)-1|\geq \zeta,\\0 \text{ otherwise.} \end{array}\right.
\end{equation}
$\zeta\in \mathbb{R}^+$ is a margin of acceptance for values close to $1$ restricting then the range of expansion and contraction allowed. $\xi\in \mathbb{R}^+$ balances the two terms. The first component heavily penalizes the negative values of $|J_\varphi(x)|$ and thus prevents violations of topology preservation and overlapping while the second one penalizes large expansions to achieve more realistic deformations.

However, this penalization does not guarantee that $|J_\varphi(x)|$ remains positive everywhere. Since the deformations must remain injective to prohibit self-penetration and distortions, we propose to add a regridding method~\cite{christensen}. This technique has the advantage of being easy to implement and being performed simultaneously with the resolution without slowing the computations. The idea is to reinitialize the registration process as soon as $|J_\varphi(x)|$ becomes too small by taking as the new moving image the previous computed deformed image. The final deformation is the composition of all the deformations. The algorithm is summarized next:

\begin{shaded*}
\vspace{-0.5cm}
\begin{lstlisting}[language=Python, mathescape=true]
#pseudocode for computing Regridding Step
Initialization: $\varphi=Id$, $regrid\_count=0$.
for $t=1,\cdots$:
    if $|J_{\varphi^{t+1}}(x)|<tol$:
        $regrid\_count=regrid\_count+1$
         $\overline{f}_{w-z+1}(x)=\overline{f}_{w-z+1}(\varphi^t(x))$
         set $tab\_phi(regrid\_count)=\varphi^t$.
         $\varphi^{t+1}=Id$
$\varphi^{final}=tab\_phi(1)\circ \cdots \circ tab\_phi(regrid\_count)$.
\end{lstlisting}
\vspace{-0.5cm}
\end{shaded*}

Eventually, the deformation model plays a crucial role in the accuracy and the speed of the method, and defines the representation of $\varphi$. In order to reduce the computational cost, we consider free-form deformations~\cite{rueckert99,sederberg86}, in which a rectangular lattice, initialized as uniformly spaced points, is superimposed on the image pixel grid and is deformed while the deformation on the finer pixel grid is recovered using a summation of tensor of cubic B-splines for their local support and smoothness. We therefore parameterize $\varphi$ as follows:
\begin{equation}
    \varphi(x)=\underset{j_1=0}{\overset{3}{\sum}}\underset{j_2=0}{\overset{3}{\sum}}P_{j_1j_2}\underset{k=1}{\overset{2}{\Pi}}\lambda_{j_k}(x_k),
\end{equation}
with $x=(x_1,x_2)\in \mathbb{R}^2$, $\{\lambda_0(x) = \frac{(x-1)^3}{6},\,\lambda_1(x)=\frac{(4-6x^2+3x^3)}{6},\\\lambda_2(x)= \frac{1+3x+3x^2-3x^3}{6},\,\lambda_3(x)=\frac{x^3}{6}\}$ are the basis spline functions, and $P_{j_1j_2}$ are the control points constituting the lattice. With this formulation, explicit expressions of the derivatives and of $|J_\varphi(x)|$ can be easily derived and the problem amounts to find the control points to reconstruct the deformation field $\varphi$.

To achieve small computational cost with accurate results, we use the Levenberg-Marquardt optimization method to solve this problem in a multi-level setting.

The final stage of our approach consists of finding the best approximation for the damaged regions from $f_{w}^{align}$. This process is achieved by iteratively minimizing patch distances in the form:
\begin{equation}
\begin{aligned}
\label{eq:EckhardYoung}
     &f_{w}^{out}(y,\Phi)= \sum_{y\in \overline{\mathcal{G}}} \mathfrak{F}(f_{1}(y),f_{0}(y,\Phi))
\end{aligned}
\end{equation}

\noindent
where $\mathfrak{F}$ is a distance measure and $\Phi$ is the shift-map between the damage and search space pixels.



\section{Experimental Results}
This section describes the set of experiments that we conducted to evaluate our solution.

\begin{figure}[!t]
\centering
\includegraphics[width=0.95\textwidth]{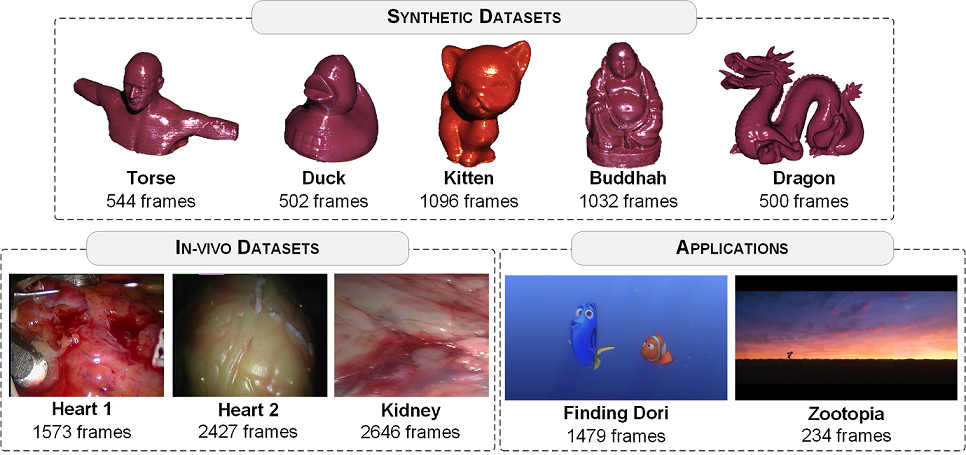}
\caption{Sample frames of the different datasets we used to evaluate our proposed solution.}
\label{fig::Datasets}
\end{figure}

\subsection{Data Description} \label{sub:DataDesc}
We evaluate our approach using ten datasets coming from different sources:  \medskip

\noindent
\textbf{Vision from Graphics Data}. We generated five datasets for our experiments, in which four are from synthetic-to-physical datasets, and one is rendered. Following common practice for graphics data (e.g.,~\cite{Netz::13}), for the first group we used 3D object models from the AIM@SHAPE repository~\footnote{http://visionair.ge.imati.cnr.it}
that we texturized, 3D printed and painted with red glossy paint. We then created the video sequences by positioning the objects on a moving station with a non-fixed source of light, and with different types of lighting. For the remaining dataset, we created video sequences directly from the 3D model with rotational motion of the object and change of lighting. Our main motivation to follow this procedure is that we can generate the specular-free versions for quantitative analysis purposes.

\medskip
\noindent
\textbf{Medical Imaging Data}. The medical domain is a very challenging scenario, in which the problem of removing specular reflections from the video sequences is of great interest. It is indeed a common preliminary stage in medical image processing followed by subsequent tasks such as tracking, stereo reconstruction and segmentation whose robustness and accuracy heavily depend on having a consistent surface appearance~\cite{Stoyanov::10}. In this setting, one can find large and very small damage areas, moving camera and non-fixed light conditions (significant changes on illumination conditions over time). Therefore, to test the robustness under these conditions, we use three in-vivo datasets that come from endoscopic video sequences~\footnote{http://hamlyn.doc.ic.ac.uk/vision/}.

\medskip
\noindent
\textbf{Entertainment Data}.
To demonstrate the effectiveness and generalization ability of our approach, we evaluate our method on two datasets from the entertainment industry. We obtained two video sequences from popular movies with objects-removal task in mind. For both datasets, we created masks for the objects to be removed from the scene and used that to evaluate the performance of the inpainting part of our approach on standard  entertainment data.

Further details on the datasets can be seen in Figure~\ref{fig::Datasets}. All results and comparisons were run under the same condition using an Intel(R) Core i7- 6700  at 3.40GHz-64GB RAM, and a Nvidia GeForce GT 610.

\subsection{Evaluation Scheme.}\label{sub:EvalSch}
To validate our approach, we design a two-part evaluation scheme, where the protocol for each part is as follows: \smallskip

\textbf{E1. Specular Reflection Detection.} To demonstrate the advantages of our algorithmic approach, we firstly offer visual comparison of our method against: TAN05\cite{Tan::05}, SH09\cite{Shen::09}, AK15\cite{Akashi::14}, and YAM17\cite{Yamamoto::17} in which the philosophy is closely related to ours. To further support the visual results, we performed a quantitative analysis which is based on the Dice's coefficient along with the CPU time for two datasets with ground truth annotations. 

\textbf{E2. Video Recovery Approach}. To evaluate the global performance of our approach, we compared it against one of the state-of-the-art methods in the area NW14~\cite{Newson::14}. Although the approaches of~\cite{Wexler::04, Patwardhan::05, GranadosB::12} are indeed interesting, they rely on a specific modeling hypothesis such as static background, color distances and homographic transformations that somehow make comparisons unfair. However, NW14~\cite{Newson::14} demonstrates a more general approach. With this motivation in mind, in this work we offer a detailed comparison against NW14~\cite{Newson::14}. Finally, we use a case study where our approach can be applied - the task of object removal.

\subsection{Results and Discussion}
In this section, we describe our findings following the scheme described in subsection \ref{sub:EvalSch}.

\begin{figure*}[!t]
\centering
\includegraphics[width=0.95\textwidth]{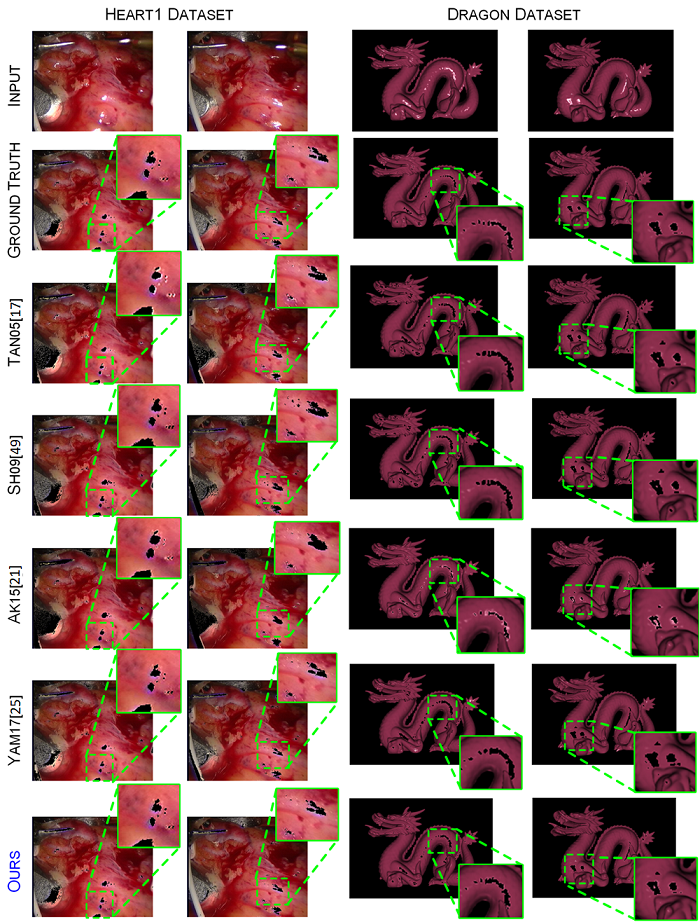}
\caption{Labeling results of our approach compared against four state-of-the art methods on two of the datsets.}
\label{fig::DetectionResults}
\end{figure*}

\begin{table}[]
\centering
\caption{Quantitative evaluation of our detection process.}
\label{tab::labelingApproach}
\scalebox{0.9}{
\begin{tabular}{lcccc}
\cline{2-5}
\multicolumn{1}{l|}{} & \multicolumn{4}{c|}{Evaluaton Measure (average)} \\ \cline{2-5}
 & \cellcolor[HTML]{EFEFEF}\begin{tabular}[c]{@{}c@{}}Dice's coefficient\end{tabular} & \multicolumn{1}{l}{\cellcolor[HTML]{EFEFEF}Accuracy} & \cellcolor[HTML]{EFEFEF}Precision & \cellcolor[HTML]{EFEFEF}\begin{tabular}[c]{@{}c@{}}Error Rate \%\end{tabular} \\ \hline
\cellcolor[HTML]{EFEFEF}Torso & 0.7230 & 0.9978 & 0.6966 & 2.2\\ \hline
\cellcolor[HTML]{EFEFEF}Kitten & 0.8214 & 0.9983 & 0.8304 & 1.17 \\ \hline
\cellcolor[HTML]{EFEFEF}Duck & 0.8741 & 0.998 & 0.835 & 1.15 \\ \hline
\cellcolor[HTML]{EFEFEF}Buddah & 0.7039 & 0.9969 & 0.6642 & 3.1 \\ \hline
\cellcolor[HTML]{EFEFEF}Heart 1 & 0.8179 & 0.994 & 0.9883 & 2.6 \\ \hline
\cellcolor[HTML]{EFEFEF}Heart 2 & 0.9175 & 0.9985 & 0.9527 & 1.5 \\ \hline
\cellcolor[HTML]{EFEFEF}Kidney & 0.8771 & 0.9981 & 0.9535 & 1.9 \\ \hline
\end{tabular}
}
\end{table}


\begin{table}[]
\centering
\label{tab::Compare}
\caption{Performance comparison against state-of-the-art methods. Auto denotes if parameters selection is automatic or manual.}
\scalebox{0.87}{
\begin{tabular}{lcc|cc|c}
                                               & \multicolumn{2}{c|}{\cellcolor[HTML]{EFEFEF}Dice's Coefficient} & \multicolumn{2}{c|}{\cellcolor[HTML]{EFEFEF}CPU Time}          & \cellcolor[HTML]{EFEFEF}                                                                                 \\ \cline{2-5}
                                               & \cellcolor[HTML]{EFEFEF}Heart  & \cellcolor[HTML]{EFEFEF}Dragon & \cellcolor[HTML]{EFEFEF}Heart & \cellcolor[HTML]{EFEFEF}Dragon & \multirow{-2}{*}{\cellcolor[HTML]{EFEFEF}\begin{tabular}[c]{@{}c@{}}Auto\end{tabular}} \\ \hline
\cellcolor[HTML]{EFEFEF}TAN05\cite{Tan::05}      & 0.38 & 0.26 & 9.94 & 42.01 & \xmark    \\
\cellcolor[HTML]{EFEFEF}SH09~\cite{Shen::09}     & $\sim$0.1 & $\sim$0.1 &  3.4e$-2$ & 3.2e$-2$ & \xmark   \\
\cellcolor[HTML]{EFEFEF}AK14~\cite{Akashi::14}   & $\sim$0.1 & $\sim$0.2  & 34.54 & 32.87 & \xmark   \\
\cellcolor[HTML]{EFEFEF}YAM17~\cite{Yamamoto::17} & 0.68 & 0.44 & 2.6e$-2$ & 1.7e$-2$ & \xmark   \\
\cellcolor[HTML]{EFEFEF}\textbf{Ours}          & \textbf{0.95} & \textbf{0.88} & \textbf{3.4e$-3$} & \textbf{3.2e$-3$}  &  \textbf{\cmark}                                                                                                \\ \hline
\end{tabular}
}
\end{table}

\begin{table}[]
\centering
\caption{Performance evaluation of our global solution.}
\label{tab::performance}
\scalebox{0.9}{
\begin{tabular}{llclc}
 &  & \cellcolor[HTML]{EFEFEF}\begin{tabular}[c]{@{}c@{}}CPU Time {[}s/frame{]}\end{tabular} & \cellcolor[HTML]{EFEFEF}Minima & \multicolumn{1}{c}{\cellcolor[HTML]{EFEFEF}\begin{tabular}[c]{@{}c@{}}[Min, Max] \# Iterations\end{tabular}} \\ \hline
\multicolumn{1}{|l|}{} & \cellcolor[HTML]{EFEFEF}Torso & $2.8797$ & $6.634e^{-2}$ &  $[19, 28]$\\ \cline{2-5}
\multicolumn{1}{|l|}{} & \cellcolor[HTML]{EFEFEF}Kitten & $1.5549$ & $2.504e^{-3}$ & $[16, 25]$\\ \cline{2-5}
\multicolumn{1}{|l|}{} & \cellcolor[HTML]{EFEFEF}Duck & $1.6489$ & $2.293e^{-2}$  &  $[14, 26]$\\ \cline{2-5}
\multicolumn{1}{|l|}{} & \cellcolor[HTML]{EFEFEF}Buddah & $1.5083$ & $8.651e^{-2}$ &  $[14, 30 ]$\\ \cline{2-5}
\multicolumn{1}{|l|}{\multirow{-5}{*}{\rotatebox[origin=c]{90}{Full-Rank}}} & \cellcolor[HTML]{EFEFEF}Heart1 & $3.7198$ & $1.038e^{-2}$ &  $[18, 37]$ \\ \hline \hline
\multicolumn{1}{|l|}{} & \cellcolor[HTML]{EFEFEF}Torso & $0.6158$ & $1.654e^{-4}$ &  $[16, 22]$\\ \cline{2-5}
\multicolumn{1}{|l|}{} & \cellcolor[HTML]{EFEFEF}Kitten & $0.1786$ & $3.742e^{-5}$ & $[10, 17]$\\ \cline{2-5}
\multicolumn{1}{|l|}{} & \cellcolor[HTML]{EFEFEF}Duck & $0.2433$ & $1.351e^{-3}$  &  $[12, 21]$\\ \cline{2-5}
\multicolumn{1}{|l|}{} & \cellcolor[HTML]{EFEFEF}Buddah & $0.4354$ & $4.260e^{-4}$ &  $[14, 24]$\\ \cline{2-5}
\multicolumn{1}{|l|}{\multirow{-5}{*}{\rotatebox[origin=c]{90}{Low-Rank}}} & \cellcolor[HTML]{EFEFEF}Heart1 & $1.2002$ & $7.185e^{-3}$ &  $[15, 31]$  \\ \hline
\end{tabular}
}
\end{table}

\medskip
\textbf{E1. Detection Approach.} We start by evaluating our detection approach using four performance metrics which can be seen in Table~\ref{tab::labelingApproach}. These numerical results were obtained using both the output of our detection solution and the ground truth of the corresponding dataset.
Our quantitative analysis starts by computing the Dice's coefficient ($Dc$) to measure how similar the detection result is to the ground truth. We can see that $Dc$ is greater than $0.7$ for all datasets and with an overall average of $0.82$. In terms of accuracy, we reported values greater than $0.99$ for all datasets while in terms of precision our approach ranges from $0.69$ to $0.98$ with overall average of $0.85$. As it can be seen in the last column of Table~\ref{tab::labelingApproach}, the error rate of our detection approach ranges from $1.1$ to $3.1$ with a global average error of $1.94\%$.

We then evaluate our approach by comparing its performance against TAN05\cite{Tan::05}, SH09\cite{Shen::09}, AK15\cite{Akashi::14}, and YAM17\cite{Yamamoto::17}. It is important to note that all the compared methods had built in fixed parameters that we had to adjust manually to achieve the reported output. Not doing this manual adjustment would result in a massive over-segmentation of the specular region.
We ran our detection approach and Table~\ref{tab::Compare} shows the performance comparison on two different datasets. We chose to report Dice's coefficient as it is the best metric to indicate how close the results are from the ground truth. 

In terms of computational time, unlike our proposed method, all the compared approaches have at least one iterative operation to achieve the detection. As such, they reported high computational time while we achieved the lowest computational complexity with a minimum of 3.2e$-$3 second/frame.

This performance improvement, both in accuracy and speed, is due to the fact that our method is designed to have a simple yet effective way to isolate specular regions. This is combined with the fact that our approach is able to automatically adapt and adjust to the specifications of each frame in process taking into account the changes in illumination in the scene.

This is further supported by Figure~\ref{fig::DetectionResults} where we show the output samples of the ground truth and all compared approaches on the same two datasets. Visual inspection of the results shows that outputs generated by our approach agrees with the theory, in which a spatially adaptive detection is more robust than fixed-parameter detection approaches. The zoom in views show that our approach yielded results that are more consistent with the ground truth in which specular reflection were precisely detected. Other methods failed to obtain reliable results across all frames as they either over/under segment the specular region.

\begin{figure*}[!t]
\centering
\includegraphics[width=0.99\textwidth]{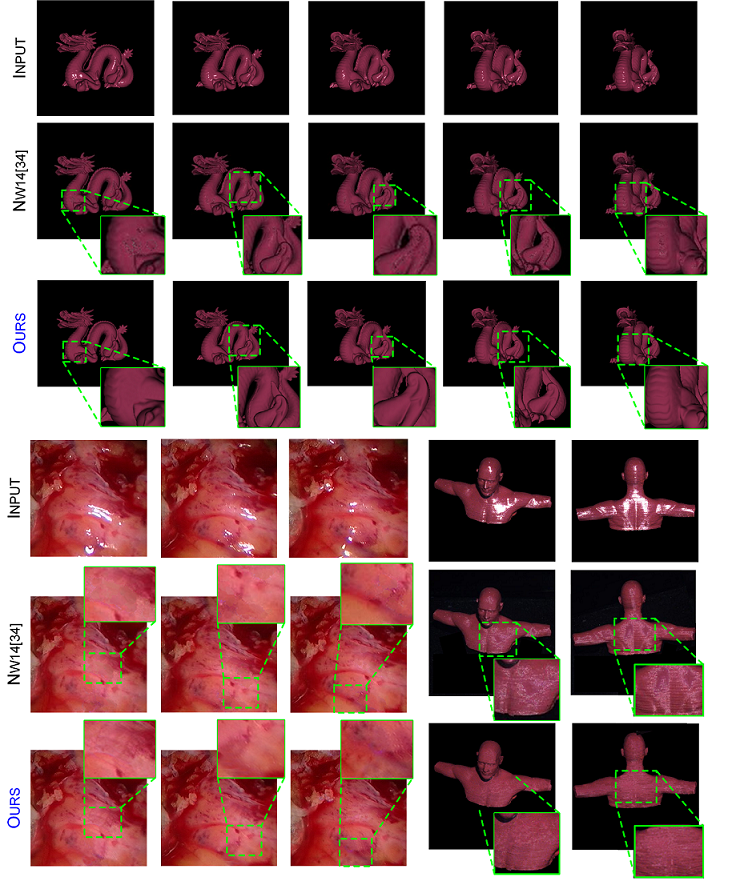}
\caption{Specular-free results of our approach compared against NW14~\cite{Newson::14}.}
\label{fig::InpaintingResults}
\end{figure*}

\textbf{E2. Restoration Approach.}
We compared our approach against Newson's algorithm~\cite{Newson::14} that is, to our best knowledge, one of the most robust approaches. We begin by a visual evaluation of our method against that of NW14~\cite{Newson::14} using three of the datasets. To this purpose, in Figure~\ref{fig::InpaintingResults}, we display interesting frames containing different cases including different sizes of damage areas, light (e.g., white and incandescent), and rigid and non-rigid objects. For example, the Torso dataset shows large areas that need to be recovered while the heart dataset represents deformable objects. Another case can be seen with the dragon dataset where there is a different kind of light, incandescent light.

Although the compared approaches gave visually good results, in many cases, the method of NW14~\cite{Newson::14} tends to over-restore areas on the boundaries of target regions producing noticeable artifacts, where inside and outside of the boundaries of the damaged region are mistakenly assigned black/grey. By contrast, our approach allows for a visually more pleasing video restoration with better preservation of details and texture. Overall, the zoom in views in the figure show that our approach consistently results in a smoother inpainting of the damage area that blends nicely with the surrounding region of the frame.

\begin{figure*}[t!]
\centering
\includegraphics[width=1\textwidth]{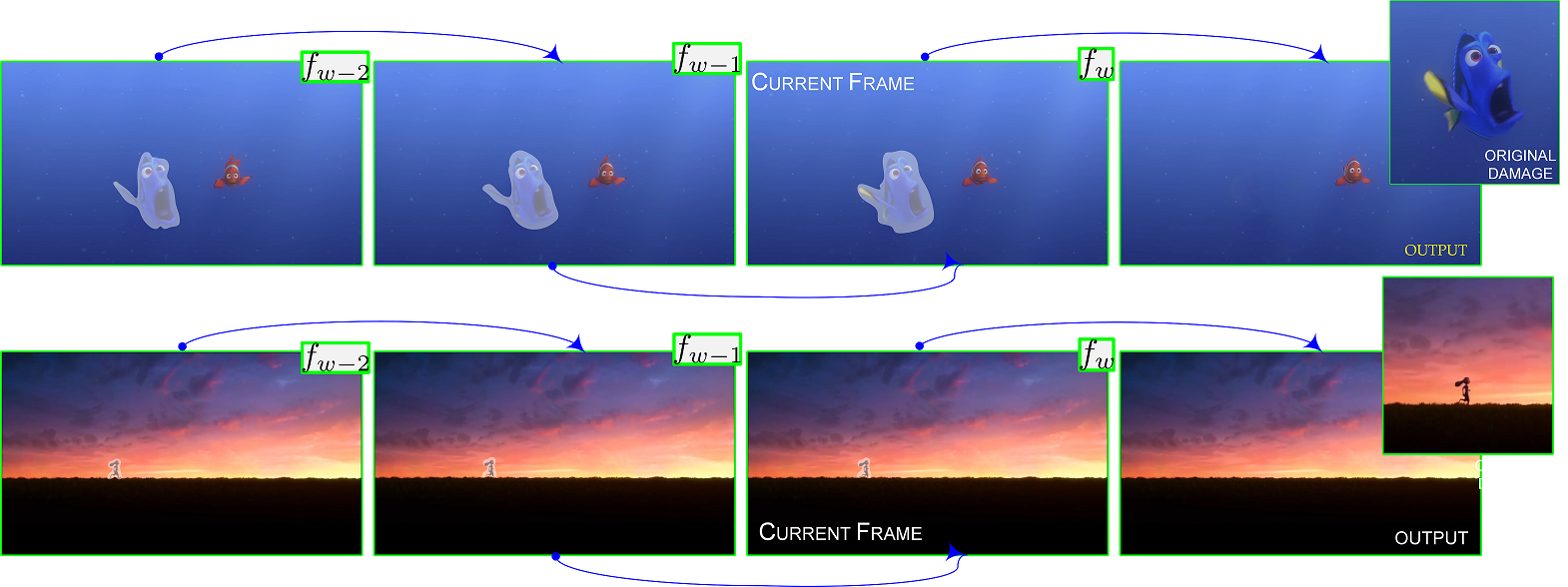}
\caption{Our approach can be used for other applications such as object removal. We display two output samples using the entertainment datasets. From left to right. A set of temporal past frames are used to capture the target object (see white highlighted objects) to be removed and to gain temporal consistency. Last column displays the result of our algorithmic approach, in which we successfully removed both \textit{Dori} and \textit{Rabbit}. 
}
\label{fig::whereIsDori}
\end{figure*}

In terms of computational time, Table~\ref{tab::performance} shows evaluation of the performance of our approach. We first want to point out the repercussion of promoting low-rank on our solution in terms of computational time. We can see that promoting low rank decreased the computational time to an average of 6 times less than the time given if using full-rank data. Moreover, the last column of the table also shows that using low-rank allows our solution to reach a better minimum with less number of iterations. On the average, full-rank demanded 23 iterations per frame while low-rank only needed 18 iterations per frame.

We ran our approach and that of NW14~\cite{Newson::14} on the same computer and under similar conditions and our approach required an average $\thicksim$0.69 sec/frame while NW14~\cite{Newson::14} needed an average of $\thicksim$33.155 sec/frame to perform the restoration process.

\textbf{Further Applications: Finding Dori.} To show the generalization capabilities of our approach, we show how it can be used for applications such as object removal. Visual results are displayed on Figure~\ref{fig::whereIsDori}, where on closer inspection one can see that the objects in mind, Dori and Rabbit, are successfully removed offering pleasing visual results.

\section{Conclusion}
In this work, we addressed the challenging problem of specular-free video recovery. We proposed a new framework, in which two contributions are introduced. The first is a spatially adaptive detection approach that searches for specular regions allowing for a better bounding of the damage areas. The second is a variational based solution for restoring efficiently the damage areas that exploits spatio-temporal correlations by representing prior data in a low-rank manner.
We showed that this combination allows for an improvement with respect to the state-of-the-art in terms of reducing over-restoration yielding to visually more pleasing results with less artifacts. Finally, we show that our work can be applied to other tasks such as object removal.


\bibliography{main}

\end{document}